\documentclass{edm_article}
\usepackage[table]{xcolor}
\usepackage{booktabs}
\usepackage{multirow}
\usepackage{amsmath}
\usepackage{graphicx}

\usepackage{multirow}
\usepackage{makecell} 
\usepackage[linesnumbered,ruled,vlined]{algorithm2e}
\SetKwBlock{DoParallel}{do $l \gets 1$ \KwTo $L$ in parallel}{end}
\usepackage{hyperref}
\usepackage{soul}
\usepackage{tabularx}
\usepackage{longtable}
\usepackage{accessibility} 
\usepackage{tikz}
\usepackage{xcolor}
\definecolor{keywords}{RGB}{0,0,255}
\definecolor{comments}{RGB}{0,128,0}
\definecolor{strings}{RGB}{255,0,0}
\usepackage{listings}
\usepackage{bbm}
\usepackage{bbm}
\usepackage{ulem, soul}
\usepackage{colortbl} 
\usepackage{natbib}
\usepackage{booktabs}
\usepackage{colortbl}
\definecolor{lightgray}{HTML}{EFEFEF}
\usepackage{ulem}
\usepackage[most]{tcolorbox}
\tcbset{
    mybox/.style={
        colback=gray!10,    
        colframe=gray!50,   
        fonttitle=\bfseries, 
        boxrule=0.4mm,      
        title=#1            
    }
}

\begin{document}

\title{A LLM-Powered Automatic Grading Framework with Human-Level Guidelines Optimization}


%
%
%
%

\numberofauthors{11} 
\author{
\alignauthor
Yucheng Chu \titlenote{Both authors contribute equally.}\\
       \affaddr{Michigan State University}\\
       \email{chuyuch2@msu.edu}
\alignauthor
Hang Li\\
        \affaddr{Michigan State University}\\
       \email{lihang4@msu.edu}
\alignauthor Kaiqi Yang\\
        \affaddr{Michigan State University}\\
       \email{kqyang@msu.edu}
\and  
\alignauthor Harry Shomer\\
        \affaddr{Michigan State University}\\
       \email{shomerha@msu.edu}
\alignauthor Yasemin	Copur-Gencturk\\
       \affaddr{University of Southern California}\\
       \email{copurgen@usc.edu}
\alignauthor Leonora Kaldaras\\
       \affaddr{Texas Tech University}\\
       \email{leonora.kaldaras@ttu.edu}
\and
\alignauthor Kevin Haudek\\
       \affaddr{Michigan State University}\\
       \email{haudekke@msu.edu}
\alignauthor Joseph	Krajcik\\
       \affaddr{Michigan State University}\\
       \email{krajcik@msu.edu}
\alignauthor Namsoo Shin\\
       \affaddr{Michigan State University}\\
       \email{namsoo@msu.edu}
\and
\alignauthor Hui Liu\\
       \affaddr{Michigan State University}\\
       \email{liuhui7@msu.edu}
\alignauthor Jiliang Tang\\
       \affaddr{Michigan State University}\\
       \email{tangjili@msu.edu}
}


\maketitle

\begin{abstract}

    Open-text responses provide researchers and educators with rich, nuanced insights that multiple-choice questions cannot capture. When reliably assessed, such responses have the potential to enhance teaching and learning. However, scaling and consistently capturing these nuances remain significant challenges, limiting the widespread use of open-text questions in educational research and assessments.
    In this paper, we introduce and evaluate \textit{GradeOpt}, a unified multi-agent automatic short-answer grading (ASAG) framework that leverages large language models (LLMs) as graders for short-answer responses. More importantly, \textit{GradeOpt} incorporates two additional LLM-based agents—the \textit{reflector} and the \textit{refiner}—into the multi-agent system. This enables \textit{GradeOpt} to automatically optimize the original grading guidelines by performing self-reflection on its errors.
    To assess \textit{GradeOpt}’s effectiveness, we conducted experiments on two representative ASAG datasets, which include items designed to capture key aspects of teachers’ pedagogical knowledge and students' learning progress. Our results demonstrate that \textit{GradeOpt} consistently outperforms representative baselines in both grading accuracy and alignment with human evaluators across different knowledge domains. Finally, comprehensive ablation studies validate the contributions of \textit{GradeOpt}’s individual components, confirming their impact on overall performance.


\end{abstract}



\section{Introduction}
\label{sec:introduction}


Accurate evaluation of assignments and examinations in a timely manner is vital to learning due to the significance of performance measurement in the learning process \cite{suzen2020textmining}. 
Traditionally, multiple-choice questions (MCQs), which asks students to select the correct answer from distracting options, dominated learning assessment studies. While this approach makes data available promptly ~\citep{mohler2011learning, burrows2015eras}, it falls short in proving insights into learners' thinking. Open-ended short-answer questions (SAQs) can provide deeper insights into students' answering rationale and knowledge concepts. This is because they are known to elicit the thinking path that describes how a student arrives at their conclusion~\citep{leacock2003c-rater}. Unfortunately, grading open-ended textual answers is tedious as substantial resources and time are needed to train raters to accurately and consistently code responses~\citep{ roy2015perspective}. More importantly, the inconsistent or unfair assessments, caused by diverged interpretations, biases, or mistakes create another challenge to SAQs grading in practice~\citep{suzen2020textmining}. To mitigate these issues and provide timely and consistent evaluation, automatic short-answer grading (ASAG) \cite{burrows2015eras} systems have become appealing. ASAG, which can be traced back to the 1960s, has bloomed in recent years due to advancements in natural language processing (NLP)~\cite{leacock2003c-rater, xie2024grade}. Early ASAG systems often used pattern-matching techniques and hand-crafted features~\cite{leacock2003c-rater}. Thus, those systems required intensive human labor to build and were limited to a few specific grading tasks. The rise of deep learning (DL) has lessened the amount of burdensome feature designs needed for early ASAG systems. DL provides an end-to-end solution that automatically learns to output grading scores from a large number of graded answer samples~\cite{ hassan2018automatic}. Due to the strong data-fitting capability of DL models, DL-based ASAG systems are able to be extended to different tasks if a large number of annotated samples are available. However, when the annotated sample size is limited, DL-based ASAG systems often face serious over-fitting issues. Beyond that,
as DL is a black-box model whose results lack interpretation, the application of DL-based ASAG systems is still limited~\cite{condor2024explainable}. 

The emergence of pre-trained language models (PLMs) and the more advanced Large Language Models (LLMs) have recently revolutionized the design of ASAG systems due to their human-like language ability and human-interpretable intermediate textual results. Therefore, many recent studies have attempted to build ASAG systems with LLMs. Promising results have been demonstrated that using fine-tuning \cite{latif2023finetuning} and prompting techniques such as Chain-of-Thought (CoT) \cite{cohn2024science} and in-context learning \cite{lee2024applyingllm}. Yet these recent techniques are still limited due to LLMs' inherent limitations such as sensitivity to prompts, context window restriction, etc., making the complex ASAG task challenging for the LLM grader. In reality, accurate, standardized, and unambiguous guidelines are critical to help human graders formulate a precise interpretation of scoring criteria.  For LLM-based ASAG systems, those guidelines also serve as the principal instructions. They teach LLMs to perform the grading task following a similar standard as human graders. However, using guidelines composed by pedagogical experts directly for LLMs is sub-optimal since the general-purposed LLMs lack domain-specific knowledge and can misinterpret the guidelines~\citep{leacock2003c-rater}. Meanwhile, LLMs are often sensitive to various facets of prompts~\citep{jiang2020knowlmsknow} where minor changes could lead to great differences in LLM’s performance. Optimizing the guidelines manually for LLMs can further take a lot of trial and error. Thus, recent works propose to conduct guideline modification with LLMs to offload human burden~\cite{cohn2024science}. While the modified guidelines yield performance improvement, the prompt search space in these methods is relatively limited. Because of this, the modified guidelines are not necessarily optimal. Additionally, abundant human efforts such as timely feedback or a large amount of labeling are required. Therefore, methods to optimize grading guidelines automatically and effectively are still desired.

In this paper, we propose a unified \textbf{multi-agent} ASAG framework that \textbf{automatically optimizes} grading guidelines. Specifically, it employs an iterative reflection mechanism to generate task prompts (guidelines) that effectively capture learners' thinking and knowledge from a small dataset of short answers. To achieve this, we innovatively introduce prompt optimization in ASAG, framing grading guideline refinement as an optimization problem aimed at maximizing accuracy. Inspired by APO \cite{pryzant2023gradient}, we develop novel techniques such as \textit{misconfidence}-based selection, iterative optimization, and log-probability-based robustness to enhance the framework’s \textbf{stability} in producing accurate and \textbf{trustworthy} score predictions on unseen datasets. To minimize human labeling effort, our mechanism intelligently selects short-answer samples that contribute to optimal guideline refinement. Additionally, the framework supports assessments across varying levels of complexity, offering interpretable evaluations for each learning objective while improving overall scoring accuracy. To validate our approach, we conducted experiments on two real-world grading datasets. The first dataset comprises responses from high school students within a physical sciences curriculum, while the second consists of a national sample of teachers answering questions designed to assess content-specific knowledge required for teaching~\citep{copur2022mathematics}. Experimental results demonstrate that \textit{GradeOpt} outperforms representative baselines in both accuracy and alignment. Further analysis highlights consistent improvements in test accuracy across iterations, showcasing the framework’s ability to continuously enhance grading guidelines. To the best of our knowledge, we are the first to apply prompt optimization in ASAG by refining grading guidelines akin to generating an optimal task prompt. We believe that our multi-agent reflective mechanism can unlock the full potential of LLMs in learning analytics by providing detailed and accurate assessments while significantly reducing educators' grading workload.

\begin{figure*}[htbp]
\Description{Illustration of the proposed framework}  
\centering
\includegraphics[width=0.95\textwidth]{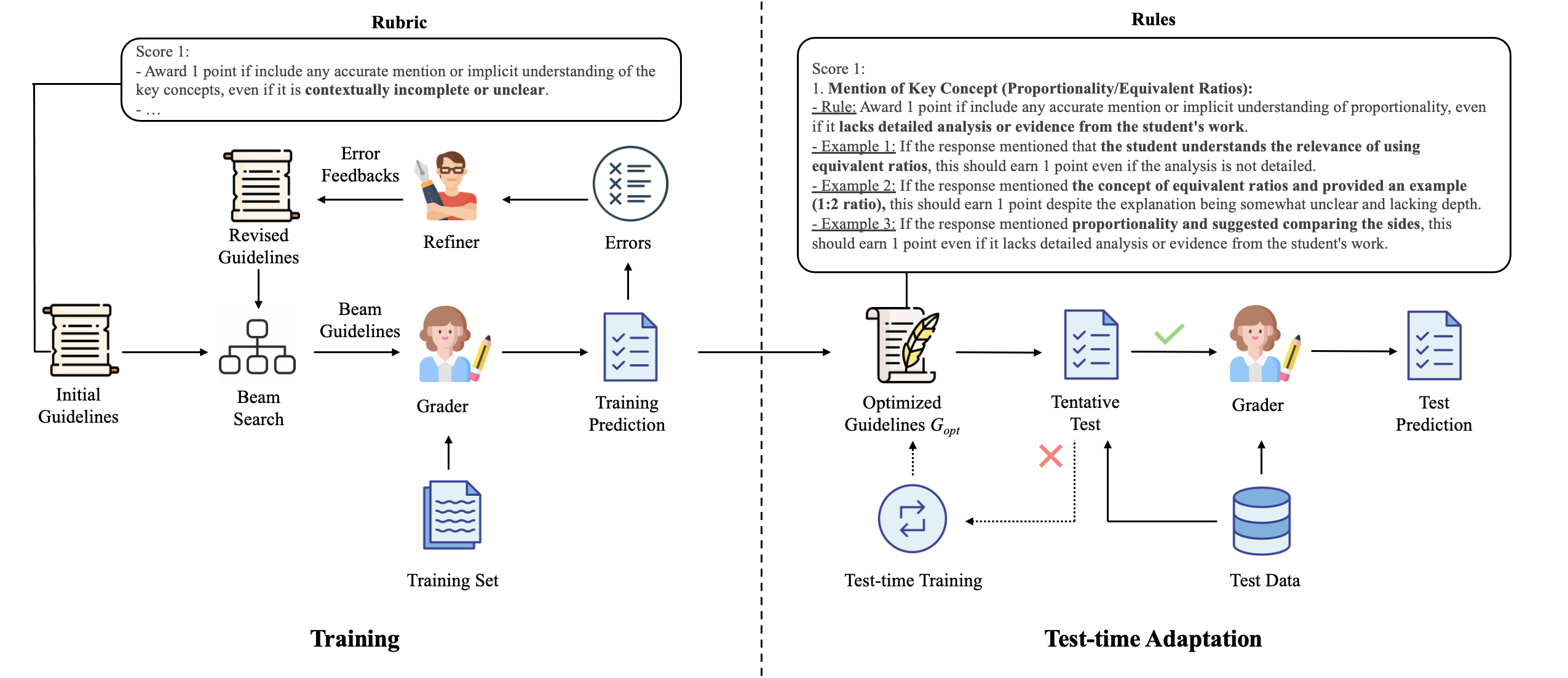} 

\caption{An Illustration of the proposed framework.}
\label{fig:framework_illustration}
\end{figure*}

\section{Related Work}

\subsection{Automatic Short Answer Grading}

Automatic Short Answer Grading (ASAG) is often treated as a text classification or regression problem in NLP studies. Here we mainly focus on classification due to its relevancy to our setting. Traditional ASAG models mainly rely on text similarity and employ classic ML classifiers. They use lexical features such as bag-of-words (BOW) \cite{mohler2011learning} and TF-IDF \cite{del2023gradeaid}, or syntactic features indicating the structure of sentences~\cite{leacock2003c-rater}. However, these methods require significant manual design, which makes them hard to be applied to new datasets. To reduce the burden of feature engineering, Deep Neural Networks (DNNs) such as Long-Short-Term-Memory (LSTM) are utilized \cite{hassan2018automatic}, which produce superior results but suffer from limited generalizability. Pre-trained BERT-based models provide enhanced versatility through \textit{transfer learning} including on ASAG datasets \cite{camus2020investigating}. To further enhance grading accuracy, researchers have made attempts to ensemble BERT with statistics-based methods ~\cite{erickson2020automated} and data augmentation \cite{lun2020multiple}. LLMs are increasingly utilized in ASAG and similar assessment tasks~\cite{yang2024content,cohn2024science,chu2025enhancing}. However, their prompts are mostly always manually-crafted and thus are unable to properly adapt to new datasets. To solve this issue, several works have shifted attention to assisting educators with guideline creation~\cite{xie2024grade,cohn2024science}.

\subsection{LLM Prompt Optimization and Reflection}

Prompts are critical to the success of LLMs~\cite{zhou2023ape}. To tailor LLMs to challenging tasks, manually crafted prompts are adopted to enhance the performance~\cite{wei2023cot}. To automate the generation and optimization of prompts, prompt optimization emerges as a promising method for input prompt refinement. 

Using these techniques, LLMs have demonstrated superior performance in many down-stream tasks, particularly in instruction following and reasoning \cite{pryzant2023gradient, zhou2023ape, yang2024large}. However, such automatic methods are risky when directly applied to ASAG tasks considering the limitations of LLMs such as hallucination~\cite{huang2023surveyhallucinationllm} and misalignment~\cite{kenton2021alignment}. To enhance both accuracy and trustworthiness, we adopt the idea of state-of-the-art prompt optimization APO \cite{pryzant2023gradient} and implement novel techniques for reliability. Similar to how humans gather knowledge from failures, experience-and-reflect~\cite{pan2023automatically} is an important technique for improving LLMs' alignment with task specifications. By reflection, LLMs learn through failure, which enriches its knowledge base and provides valuable reference in similar scenarios. Self-reflection has demonstrated promising results in improving LLM reasoning  \cite{shinn2023reflexion, madaan2023selfrefine}. However, LLMs' reflection ability is relatively limited when it comes to self-correction without human feedback or true labels \cite{huang2024largelanguagemodelsselfcorrect}. A recent work \cite{tyen2024llmscannot} divides the task of self-correction into two steps: mistake finding and output correction. They empirically show that while LLMs struggle to find errors, their correction ability is robust when given ground-truth labels. This provides grounding support for our proposed framework due to the similar use of true labels in guiding LLM reflection.

\section{Problem Statement}
We define ASAG as a text classification task, which grades the short answer text $x_i$ by classifying it into the discrete score categories $\{\hat{y}_i=\mathcal{F}(x_i)=s_j|j=1,..,C\}$, where $\mathcal{F}$ is a ASAG system, $\hat{y}_i$ is the score prediction, $s_j$ is the score category, and $C$ is size of the score category set. When $\mathcal{F}$ is an LLM, the grading guideline text $G$ will be concatenated at the front of $x_i$ as an instructional prompt, and the grading process can be expressed as $\hat{y}_i=\mathcal{F}(G,x_i)=s_j$. In this work, we focuses on leveraging the reflection and refining capabilities of LLMs to automatically generate an optimized grading guideline $G^*$ based on a small amount of graded short answer text $\mathcal{D}=\{(x_i,y_i)|i=1,...,N\}$, where $N$ is the number of graded samples. The goal of our framework can be expressed as: $ G^*=\underset{G}{\text{argmax}\ }\frac{\Sigma_{i=1}^{N}\mathbbm{1}_{y_i=\hat{y}_i}}{N},\ G \in \mathcal{G}$,where $\mathcal{G}$ is the potential grading space, $\mathbbm{1}_{\{\cdot\}}$ is an indicator function that is 1 if $y_i=\hat{y}_i$ and 0 otherwise. Once the optimization process is finished, our framework will concatenate the optimized guidelines $G^*$ at the front of unlabeled short answer text and generate the grading results, $\hat{y}=\mathcal{F}(G^*,x_i)$.

\section{Method}\label{framework}

In this section, we introduce our unified multi-agent ASAG framework \textit{GradeOpt}.  It can automatically optimize the grading guidelines and achieve better grading alignment with human experts. Next, we first give an overview of \textit{GradeOpt}. Then, we detail the LLM-based agent design, and implementation details.

\subsection{An Overview}

As demonstrated in Figure \ref{fig:framework_illustration}, \textit{GradeOpt} consists of two stages: \textbf{\textit{training}} and \textbf{\textit{test-time adaptation}}. The training stage is supported by three LLM-based agents: \textbf{\textit{Grader}}, \textbf{\textit{Reflector}}, and \textbf{\textit{Refiner}}. They synergically enhance the grading guidelines by optimizing the score classification accuracy using the graded answers to the SAQs $\mathcal{D}_{train}$ (i.e., the training data). In the test-time adaptation stage, the system first performs an out-of-distribution (OOD) test over a small amount of unlabeled answers sampled from the test data. To be specific, by checking the log likelihood score of the predicted grading results, \textit{GradeOpt} decides whether the optimized guidelines $G_{opt}$ can be applied to the test data directly. If the test failed, the current guideline is not optimal for the test data. Therefore, our framework will improve $G_{opt}$ via test-time training. If the test successes, \textit{GradeOpt} will perform the auto-grading over the whole test data automatically.

\begin{figure}[!btph]
\scriptsize
\centering
\begin{tikzpicture}
  \draw node[draw=black,fill=black!10,rounded corners,inner sep=2ex,text width=.445\textwidth] {

    \textbf{Task Description:}    
    You are GradeGPT. You assess teachers’ knowledge of students’ mathematical thinking by grading their responses to a pedagogical content knowledge question.  \newline

    \textbf{Question Stem:} 
    Based on the student's work, what is student likely to understand about the relationship between the length of the shadow and the height of the object? \newline
     
    \textbf{Key Concept:} 
    Teachers should infer that the student possibly understands that there is a proportional relationship between the height of an object and the length of its shadow. However, because the concept of halving/doubling is natural to students, it is unclear if the student understands the relationship between object height and shadow length is proportional or if they understand equivalent ratios. \newline

    \textbf{Scoring Rubrics:} 
    
    - Award 0 points if it does not address key concept ...
    
    - Award 1 point if the response includes an accurate mention or implicit understanding of the key concept ...
    
    - Award 2 points if the response offers a clear and explicit analysis of the proportional relationship between the objects ... \newline

    \textbf{Adaptation Rules:} 

    1. Mention of Key Concept:
    
    - Rule: Award 1 point if the teacher response includes any accurate mention or implicit understanding of the Key Concept (proportionality/equivalent ratios), even if it lacks detailed analysis or evidence from the student's work.
    
    - Example 1: If the response mentions that the student understands the relevance of using equivalent ratios, this should earn 1 point even if the analysis is not detailed.
    
   ...
    };
\end{tikzpicture}
\caption{An example of the optimized guidelines, $G^*$.}
\label{fig:guidelines_example}
\end{figure}

\subsection{Training Stage}
\label{sec:training}

The training stage is to optimize the guideline for the \textit{Grader} agent to achieve the optimal grading performance over the training dataset $\mathcal{D}_{train}$. \textit{GradeOpt} leverages a multi-agent framework powered by three agents which collaboratively predict scores for $\mathcal{D}_{train}$, identify errors, and suggest rule modifications to mitigate errors.

Before diving into the details of this stage, we first give a brief introduction to the three key components of a common grading guideline: Question Stem ($G_{qs}$), Key Concept ($G_{kc}$) and Scoring Rubric ($G_{sr}$). Specifically, $G_{qs}$ contains the complete question contents, $G_{kc}$ describes the test knowledge concepts, and $G_{sr}$ is the operational guidance instructing human graders how to score responses. As we previously mentioned, directly using $G_{0}=\{G_{qs}||G_{kc}||G_{sr}\}$ as grading guideline for \textit{Grader} is sub-optimal since the human-based scoring rubrics commonly lack  detailed explanations to some concepts. As a result, LLM-based grading methods could provide ambiguous judgments. To solve this issue, \textit{GradeOpt} focuses on optimizing $G$ by appending new Adaption Rules ($G_{ar}$) that provides the detailed explanations regarding reflections from failed predictions and identified errors. In Figure~\ref{fig:guidelines_example}, we present an example of optimized grading guideline $G_{opt}$. Specifically, when given the expert-designed input containing ``Task Description", ``Question Stem", ``Key Concept" and ``Scoring Rubrics", \textit{GradeOpt} automatically generates the additional descriptions in ``Adaptation Rules". These new rules help describe how to assign a grade based on answer patterns and details.

The training procedure is shown on the left sub-figure of Figure ~\ref{fig:framework_illustration}. During training,  the optimization is conducted in an iterative manner. In the $t$-th round, \textit{GradeOpt} first draws a batch of samples $b$ from $\mathcal{D}_{train}$ and sends them to the grader agent for grading. \textit{GradeOpt} compares the grades outputted by LLMs with human-annotated scores, then identifies error samples. These samples are then sent to the reflector agent for error reflections. Based on the reflections generated from those error samples, the reflector agent proposes a series of suggestions for improving $G_{t-1}$, represented by $\Delta G_{t}$. $\Delta G_{t}$ is then sent to the refiner agent, which fuses $G_{t-1}$ with $\Delta G_{t}$ and generates $G_{t}$ for the next iteration of optimization.  Next, we will introduce detailed designs of the three agents in \textit{GradeOpt}. Then, we will present the implementation details of the iterative optimization process.

\subsubsection{Agent Configurations}

\paragraph{\textbf{Grader}~\label{grader}}

The \textit{Grader} focuses on mapping $x_i$ to $s_j$ based on the given $G$. In \textit{GradeOpt}, we leverage the exceptional instruction-following capability of LLMs by using a prompt to instruct LLMs to simulate the grading process of human graders. To fully exploit the potential of LLMs, we incorporate the prompt engineering strategy Chain-of-Thought~\cite{wei2023cot}. This encourages LLMs to provide both judgment and intermediate reasoning steps in their outputs. With such design, the \textit{Grader} becomes better aligned with the human-like grading process. Meanwhile, the intermediate reasoning steps provide support for the  \textit{Reflector} to discover the potential improvements to the given guideline. The prompt for the \textit{Grader} agent is shown in Figure~\ref{fig:grade_prompt}.

\begin{figure}[!btph]
\begin{tcolorbox}[mybox={Grader Prompt}]
\scriptsize
\textbf{Task Description:}
    In this task, you perform the task of assessing teachers’ knowledge of students’ mathematical thinking by grading teacher's response to a math teaching question.

    \textbf{Question Stem:} 
    <question stem>
    
    \textbf{Key Concept:} 
    <key concept>

    \textbf{Scoring Rubrics:} 
    <scoring rubrics>

    \textbf{Adaptation Rules:} 
    <adaptation rules>

    \textbf{Output format}
    
    <score>
    
    Reasoning: <reasoning>
    
    \textbf{Output Rules}
    
    1. Replace <score> with only one integer from 0, 1, or 2.
    
    2. Replace <reasoning> with your reasoning.
    
    \textit{Let's think step by step!}
\end{tcolorbox}
\caption{An example of the prompt to \textit{Grader}.}
\label{fig:grade_prompt}
\end{figure}


\vspace{-0.17in}

\paragraph{\textbf{Reflector}~\label{reflector}}

The role of \textit{Reflector} is to propose ways to improve the current guideline $G_{t-1}$ by reflecting over the error samples returned by \textit{Grader}. To be specific, we design a two-step instruction prompt for LLMs to achieve this goal. In the first step,  LLM is instructed to analyze the individual and shared failure reasons for a set of error samples. Then, in the second step, we ask LLMs to propose suggestions that can help resolve those issues. In general, the two-step improving process is analogous to the gradient descent algorithm used by parameter optimization for machine learning algorithm~\citep{ruder2017overviewgradientdescentoptimization}. In our case, the guideline $G$ serves as the parameter of \textit{Grader} and identifying the error reason is similar to the ``gradient". Finally, proposing improving suggestions based on discovered reasons is similar to making a descent down the ``gradient" and thus optimizing $G_{t-1}$. The prompt for the \textit{Reflector} agent is shown in Figure  \ref{fig:reflect_prompt}.

\paragraph{\textbf{Refiner}~\label{refiner}}

The role of \textit{Refiner} is to generate a new guideline $G_{t}$ based on the suggestions from \textit{Reflector}. Specifically, \textit{Refiner} is asked to make modifications to the examples and illustrations to the content in $G_{ar}$. Such edits include adding, removing, or editing. Note that we keep the other components, i.e., $G_{qs}$, $G_{kc}$, $G_{sc}$ unchanged since they are composed by human experts, and any small change may distort the scoring logic away from its original design. The refined guideline can be expressed as $G_t=\{G_{qs}||G_{kc}||G_{sc}||G_{ar}\}$, where $||$ is the text concatenation operator. The prompt for the \textit{Refiner} agent is given in Figure  \ref{fig:refine_prompt}.

\subsubsection{Iterative Optimization Designs}

\vspace{-3mm}

\paragraph{\textbf{Nested Iteration}~\label{nested_iteration}}

The high complexity of test questions and grading guidelines makes it nontrivial to implement the optimization directly. Beyond that, the constraint over the input context window size of LLMs forbids it to accept all examples in $\mathcal{D}$ for processing at once. To resolve that, we propose a nested iterative optimization approach, i.e., inner and outer loop, in \textit{GradeOpt}. Specifically, during the $t$-th outer loop, \textit{GradeOpt} selects a batch of samples $b_{out}$ from $\mathcal{D}_{train}$ and sends them with $G_{t-1}$ to \textit{Grader} for grading. Then, the wrongly graded answers $e_t$ are filtered for reflections. However, due to the input context window size limitation, all errors in $e_t$ cannot be entirely processed by \textit{reflector} and \textit{refiner} simultaneously. Thus, we introduce the inner loop, which samples an inner batch $b_{in}$ from $e_t$, and updates $G_{t-1}$ with the iterative procedure.

\begin{figure}[!btph]
\begin{tcolorbox}[mybox={Reflector Prompt}]
\scriptsize
    
    You are ReflectorGPT, a helpful AI agent capable of reflecting on [adaptation rules] that is used by a classifier for a grading task. Your task is to reflect and give reasons for why [adaptation rules] have gotten the given examples in [failed examples] wrong. \newline

    The prompt contains two components: 1. [question stem], [key concept] and [scoring rubrics] (these three are given by experts and should not be modified); 2. [adaptation rules] (your task to modify).\newline

    \textbf{Important Steps For Devising Rules: } Read [failed examples]. For each one of the errors, perform the following steps:
    
    - Step 1: Explain why the classifier made the mistakes, and provide detailed, explanative analyses for why this teacher response should not be interpreted in that wrong way. 

    - Step 2: Devise or modify [adaptation rules] for each mistake to help classifier effectively avoid the mistake and classify the teacher response into the correct category (label). Make sure the devised rule is explanative, straightforward, detailed, concise, and in 1 to 3 sentences. \newline

    \textbf{Question Stem:} 
    <question stem>
    
    \textbf{Key Concept:} 
    <key concept>

    \textbf{Scoring Rubrics:} 
    <scoring rubrics>

    \textbf{Adaptation Rules:} 
    <adaptation rules> \newline

    But [adaptation rules] gets the following examples wrong:
    
    \textbf{Failed Examples:}
    <errors> \newline

    Give reasons for why [adaptation rules] could have gotten the examples wrong. 
    \textit{Let’s think step by step!}
\end{tcolorbox}
\caption{An example of the prompt to \textit{Reflector}.}
\label{fig:reflect_prompt}
\end{figure}

To accelerate the optimization process and encourage a wider exploration of all possible combinations of error samples in $b_{out}$, we integrate the beam searching strategy~\citep{freitag2017beam} within both inner and outer loops. The algorithm of the nested iteration is shown in Algorithm~\ref{algo:two-fold}. To be specific, in the $w$-th inner loop of the $t$-th outer iteration, \textit{GradeOpt} accepts guidelines beam $G_{t,w-1}=\{g_{t,w-1}^{(k)} \mid 1 \leq k \leq K\}$ from $(w-1)$-th inner iteration instead of a single guideline for refining (line 5). Then, during the inner iteration,  each $g_{t,w-1}^{(k)}$ will be sent for reflection and refinement with $L$ independently sampled inner batches $b_{in}$ in a parallel manner (line 9). After all refined guidelines for the $w$-th inner loop are finished, $G_{t,w}=\{g_{t,w}^{(l,k)} \mid 1 \leq l \leq L, 1 \leq k \leq K \}$, each new guideline $g_{t,w}^{(l,k)}$ will be tested over a hold-out validation set $\mathcal{D}_{val}$ (line 14). Meanwhile, the top-$K$ performing guidelines will be kept as $G_{t,w}$ and passed to the $(w+1)$-th inner loop. Finally, the beam output of the last iteration of inner loop $G_{t,W}$ will be sent to the $(t+1)$-th outer iteration (line 4).

While this procedure helps increase the accuracy and reliability, blindly increasing the iteration could lead to over-fitting and higher computational overheads\cite{juneja2024taskfacet}. This is particularly true for smaller datasets. To help address these challenges, we introduce an early-stopping criteria. Specifically, during the selection for top-K performed $G_{t,w}$ in the $w$-th inner loop, we record the performance metric $m_w$ of the best performed guideline. Then, in the next $(w+1)$-th inner iteration, we check if $m_{w+1}$ is improved. If $m_w$ stops improving for two consecutive inner iterations, it indicates that the current guideline is facing risks to be over-fitted, thus following inner iterations are skipped. Similarly, during the $t$-th outer iterations, if its inner iteration is terminated due to the early-stopping and $(t-1)$-th outer iteration's inner iteration is also terminated by early-stopping, the following outer iterations will also be skipped.

\vspace{-0.17in}

\paragraph{\textbf{Batch Sampling Strategies}~\label{minibatch}}

Using self-reflective approaches of LLMs to refine grading guidelines requires the exposure of similar errors in consecutive optimization iterations due to LLMs' lack of ability in generating appropriate modifications with one attempt \cite{ma2024llmsgood}. This is especially true for complicated cases involving nuance differences between score categories.
However, the randomness of batch sampling in the outer loop fails to guarantee this pre-requisite, which limits the performance of \textit{GradeOpt}. To solve this, we develop a novel sampling strategy, which leverages the \textit{misconfidence} metric ($\psi$)~\citep{xu2024misconfidence} to find challenging examples in $\mathcal{D}_{train}$. To be specific, given $x_i$ as an input to \textit{Grader} and $y_i$ as its human grading result, we calculate $ \psi_i = \frac{\max_{\hat{y_i} \neq y_i} \log{P_{LLM}(\hat{y_i} | G, x_i)} }{ \log{P_{LLM}(y_i | G, x_i)} }$, where $\hat{y_i}$ is the prediction of \textit{Grader}. The \textit{misconfidence} quantifies the discrepancy between the highest log probability of \textit{Grader}'s incorrect prediction $y_i$ and the log probability of correct prediction $\hat{y_i}$. Intuitively, the larger $\psi$ indicates that the \textit{Grader} is giving the wrong judgment with a relatively high confidence over the correct one, thereby implying that the sample is more challenging. However, calculating $\psi$ over all $x_i \in \mathcal{D}_{train}$ is computationally expensive and cannot be directly done in each iteration. To avoid introducing the additional computing cost to the current algorithm, we only calculate $\psi_i$ for samples in current iteration batch $x_i \in b_{out}$ and select the top-$C$ samples as seeds to query similar samples from $\mathcal{D}_{train}$ through embedding similarities. In this way, we simplify the selection process and ensure the consecutive appearance of the similar challenging examples between iterations. At last, to avoid the optimization being operated over the same portion of samples from $\mathcal{D}_{train}$ all the time, we only select half batch based on \textit{misconfidence}, and keep the another half as random samples. The detailed comparisons between the batch sampling strategies are presented in Section~\ref{ablation}.

\vspace{-0.1in}
\begin{algorithm}[!btph]
  \scriptsize
  \LinesNumbered\LinesNumberedHidden 
  \setlength{\algomargin}{0em} 
  \KwData{training split of Dataset $\mathcal{D}_{train}$, validation split of Dataset $\mathcal{D}_{val}$, initial guidelines $\mathcal{G}$, outer loop iteration number $T$, inner loop iteration number $W$, parallel inner batch number $L$, guidelines beam size $K$.}
  \KwResult{Optimized guidelines $G_{opt}$.}
  Initialize $G_{0,W}=\{g^{(k)}_{0,W}\}=\{\mathcal{G}\}$\;
  \For{$t\gets1$ \KwTo $T$}{
    $b_{out} \gets$ sample an outer iteration batch from $\mathcal{D}_{train}$ \;
    Initialize $G_{t,0}$ = $G_{t-1,W}$ \;
    \For{$w\gets1$ \KwTo $W$}{
      
      \For{$k\gets1$ \KwTo $K$}{
        $\hat{y}_{out} \gets$ generate grading results for $b_{out}$ by \textit{Grader} with guideline $g^{(k)}_{t,w}$ \;
        $e_{t,k} \gets$ find error graded samples from $b_{out}$ caused by guideline $g^{(k)}_{t,w}$ \;
        \DoParallel{
          $b_{in}^{(l)} \gets$ randomly sample an inner batch from $e_{t,k}$ \;
          $g_{t,w}^{(k,l)} \gets$ generate optimized guideline with inputting $b_{in}^{(l)}$ and $g_{t,w-1}^{(k)}$ to \textit{Reflector} and \textit{Refiner}  \;
        }
      }
      $G_{t,w} = \{g_{t,w}^{(k)} \mid 1\leq k \leq K\} \gets$ select top-$K$ performed $g_{t,w}^{(k,l)}$ based on grading performance over $\mathcal{D}_{val}$ \;
    }
  }
  \caption{Nested Iterative Prompt Optimization Algorithm }
  \label{algo:two-fold}
\end{algorithm}

\subsection{Test-time Adaptation Stage}

In this stage, \textit{GradeOpt} begins to perform the automatic grading to the large scaled unlabeled responses in test data. However, due to the diversity of language expressions existing in open-ended answers and other influence factors such as geography and time that change users' expression styles, the performance of the auto-graded is not always guaranteed to be the same as during training. Such phenomenon is well-recognized as the out-of-distribution (OOD) issue in many machine learning problems~\citep{hendrycks2016baseline}. Prior work \citep{hendrycks2016baseline} has shown that capturing prediction probability statistics about correct or in-sample examples is often sufficient for detecting whether an example is in error or abnormal. Inspired by this, we compose a confidence indicator $\zeta = \frac{1}{|\mathcal{D}_{test}|}\sum_{x_i \in \mathcal{D}_{test}}\max_{j}(\log P_{LLM}(s_j|G,x_i))$, where $\log P_{LLM}(\cdot)$ denotes the log likelihood probability given by the LLM. Intuitively, the log probability reflects the confidence that \textit{Grader} gives to its graded results. By comparing $\zeta$ with the average LLM confidence scores $\mu$ on samples in $\mathcal{D}_{train}$, we can know how serious the OOD phenomenon is. Specifically, when $\zeta > \mu$, it indicates that $G$ is well-applicable to $\mathcal{D}'$. When $\zeta < \mu$, it suggests that the guideline is facing serious OOD influences, which suggests that grader may struggle to produce reliable and accurate predictions for $\mathcal{D}_{test}$.

If the test samples are deemed to be OOD, a common solution is to first compose an adaption dataset from the testing scenario. Using this adaption dataset, we then perform test-time training on the existing model. To be specific, test-time training leverages the annotation samples from $\mathcal{D}_{test}$ and fine-tunes the optimized guideline $G_{opt}$ with the same training process introduced in Section~\ref{sec:training}. Unfortunately, in the ASAG scenario, the annotation is usually expensive. Besides, it is challenging to ask pedagogical experts to provide a large amount of annotation samples to help the existing system adapt to any changes in a timely manner. To solve this issue, we propose an incremental labeling approach which checks the marginal performance changes brought by gradually increasing the size of annotation samples. By selecting the size with highest marginal gains in metrics like accuracy and Kappa, \textit{GradeOpt} only asks pedagogical experts for necessary annotations. This not only reduces the annotation work loads but also increases the adaption efficiency of the framework. Finally, when the $G_{opt}$ passes the OOD test, \textit{GradeOpt} will be leveraged to finish ASAG over all samples in $\mathcal{D}_{test}$.

\vspace{-0.1in}
\begin{figure}[!btph]
\begin{tcolorbox}[mybox={Refiner Prompt}]
\scriptsize
    
    I'm trying to write a classifier for a grading task. You are RefinerGPT, a helpful AI agent capable of refining [adaptation rules] to be used by the classifier. \newline

    The [adaptation rules] must contain patterns learned from failed examples, explaining why the predicted score is wrong comparing to the correct label. The set of rules must strictly abide by [scoring rubrics] and must clearly use patterns/details from examples to clearly illustrate and explain. 
    
    \textbf{Question Stem:} 
    <question stem>
    
    \textbf{Key Concept:} 
    <key concept>

    \textbf{Scoring Rubrics:} 
    <scoring rubrics>

    \textbf{Adaptation Rules:} 
    <adaptation rules> \newline

    But [adaptation rules] have gotten several examples wrong, with the reasons of the problems examined as follows:

    \textbf{Failed Examples:} 
    <errors>

    \textbf{Error Feedbacks:}
    <error feedbacks> \newline

    Based on the above information, I wrote \textbf{one} different improved set of rules in replacement of [adaptation rules] for instructing the classifier to learn patterns from examples for avoiding such errors. \newline
    
    \textit{Let’s think step by step!}
\end{tcolorbox}
\caption{An example of the prompt to \textit{Refiner}.}
\label{fig:refine_prompt}
\end{figure}

\section{Experiment}

In this section, we conduct experiments to validate the effectiveness of \textit{GradeOpt}. Through the experiments, we aim to answer the following research questions. \textbf{RQ1}: Whether the refined guidelines based on prompt optimization match or exceed the performance of human-crafted guidelines? \textbf{RQ2}: Are the optimized guidelines applicable to new datasets of the same or similar questions? \textbf{RQ3}: How does each component contribute to the overall effectiveness of the guideline optimization system?

\vspace{-.25cm}
\subsection{Datasets}

To address the research questions above, we conduct experiments using two representative datasets for SAQ grading. Unlike existing ASAG studies~\citep{dzikovska2013semeval, mohler2011learning}, which focus solely on student responses, our work extends ASAG to the grading of pedagogical answers from both students and teachers. The first dataset, $\mathcal{D}_1$, consists of teachers' responses to questions designed to assess the knowledge and skills essential for teaching mathematics~\citep{copur2022mathematics}. Since grading pedagogical answers requires a more nuanced interpretation to capture the underlying thought process, evaluating \textit{GradeOpt} on this dataset allows us to examine its performance on more complex ASAG tasks. Specifically, $\mathcal{D}_1$ includes six questions addressing different aspects of teacher knowledge, with responses labeled on a three-point scale: Bad (0), Fair (1), and Good (2). The second dataset, $\mathcal{D}_2$, evaluates \textit{GradeOpt} on student responses, aligning with prior studies. It comprises 252 high school student responses to 11 assessment items within a physical sciences curriculum. These assessments measure Learning Progress (LP)-aligned scientific text-based explanations, reflecting students' ability to apply knowledge of electrical interactions in high school Physical Science~\cite{kaldaras2021developing}. Responses in $\mathcal{D}_2$ are graded on a binary scale: Fail (0) or Pass (1). All grading labels in both $\mathcal{D}_1$ and $\mathcal{D}_2$ were assigned by at least two human raters. In cases of disagreement, a third rater provided the final judgment. Detailed statistics for both datasets are presented in Table~\ref{tab:response_count_table}. For our experiments, we split both datasets into training, validation, and test sets using a 7:1:2 ratio.

\vspace{-0.5cm}
\begin{table}[!btph]
\centering
\caption{Detailed statistics of different questions in both datasets $\mathcal{D}_1$ and $\mathcal{D}_2$. The number of samples in each label category is shown as $C_i$.}
\label{tab:response_count_table}
\renewcommand{\arraystretch}{1.07}
\resizebox{0.475\textwidth}{!}{
\begin{tabular}{@{}ccl|ccl@{}}
\toprule
\multicolumn{3}{c|}{$\mathcal{D}_1$} & \multicolumn{3}{c}{$\mathcal{D}_2$} \\ \midrule
\textbf{Question} & \textbf{Total} & \textbf{$C_1$ / $C_2$ / $C_3$} & \textbf{Question} & \textbf{Total} & \textbf{$C_1$ / $C_2$} \\ \midrule
\rowcolor[HTML]{EFEFEF}$Q_1$ & 261 & 36 / 104 / 121 & $Q_1$ & 252 & 43 / 209 \\
$Q_2$ & 265 & 78 / 47 / 140 & $Q_2$ & 252 & 123 / 129 \\
\rowcolor[HTML]{EFEFEF}$Q_3$ & 236 & 132 / 66 / 38 & $Q_3$ & 252 & 113 / 139 \\
$Q_4$ & 231 & 180 / 44 / 7 & $Q_4$ & 252 & 183 / 69 \\
\rowcolor[HTML]{EFEFEF}$Q_5$ & 232 & 83 / 112 / 37 & $Q_5$ & 252 & 242 / 10 \\
$Q_6$ & 229 & 74 / 43 / 112 & $Q_6$ & 252 & 244 / 8 \\
\rowcolor[HTML]{EFEFEF}$Q_7$ & 230 & 64 / 114 / 52 & $Q_7$ & 252 & 245 / 7 \\
$Q_8$ & 231 & 108 / 24 / 99 & $Q_8$ & 252 & 227 / 25 \\
\rowcolor[HTML]{EFEFEF}- & - & - & $Q_{9}$ & 252 & 243 / 9 \\
- & - & - & $Q_{10}$ & 252 & 210 / 42 \\
\rowcolor[HTML]{EFEFEF}- & - & - & $Q_{11}$ & 252 & 241 / 11 \\ \bottomrule
\end{tabular}}
\end{table}

\vspace{-.25cm}
\subsection{Baselines}

We compared our model with several representative ASAG baselines. Firstly, we choose two popular non-LLM methods, i.e., SBERT~\citep{reimers2019sentencebert} with Logistic Regression and RoBERTa~\citep{liu2019roberta} with Fine-tuning. Both of them have demonstrated strong performance in prior studies~\cite {condor2021automaticSA,poulton2021explaining}. In addition, we adopt GPT-4o with zero-shot prompting, referred to as GPT-4o, as another baseline. Compared with non-LLM methods, LLM's exceptional instruction and human-like reasoning capabilities make it a powerful method when facing complicated grading cases~\citep{henkel2024can}. To mitigate the manual burden of revising the guidelines in the GPT-4o setting, we implement and compare \textit{GradeOpt} with APO~\citep{pryzant2023gradient}, which is a state-of-the-art method for automatic prompt optimization tasks.

\vspace{-.25cm}
\subsection{Implementations}

To implement the nested iterative optimization, we set the outer batch size $|b_{out}| = 64$ and inner batch size $|b_{in}|=8$. The outer loop iteration number $T=5$ and the inner loop iteration number $W=3$. We implement the beam search selection mechanism with Upper Confidence Bound (UCB)~\citep{auer2003using}, where the guideline beam size $K=4$. The evaluation metric for UCB is Cohen's Kappa as it empirically works better than other metrics. The agents in our framework are all powered by GPT-4o \cite{openai2024gpt4} with zero-shot prompting. The temperature for \textit{Grader} is set to 0.0 to decrease the randomness of the result. The temperatures for both \textit{Reflector} and \textit{Refiner} are set to 0.5, since we want to encourage the LLMs to be more open in exploring the error reasons and propose the improving suggestions. For each question, we run the algorithm 3 times and report the average results.

\vspace{-.25cm}
\subsection{Evaluation Metrics}

In this work, we use Accuracy (Acc) and Cohen's Kappa ($\kappa_c$) as the evaluation metrics to compare the performance of different models. To be specific, accuracy measures the percentage of correct predictions across all cases, while Cohen's Kappa measures the inter-rater alignment between model's predictions and expert annotations, accounting for agreement by chance. For the $\mathcal{D}_1$ dataset, which involves multi-class classification, we additionally utilize Quadratic Weighted Kappa ($\kappa_w$), which is particularly suitable for ordinal data as it assigns different weights to disagreements based on their magnitude.

\vspace{-.25cm}
\subsection{Main Results} \label{sec:pilot_result}

In this section, we address \textbf{RQ1} by comparing baseline models with \textit{GradeOpt} on both datasets, $\mathcal{D}_1$ and $\mathcal{D}_2$. Table~\ref{tab:result_pilot} presents the performance of baseline models and \textit{GradeOpt} on $\mathcal{D}_1$. The results reveal several key observations. While all models achieve relatively high accuracy across questions, the Cohen’s kappa values for baseline models such as RoBERTa and SBERT on some questions are notably low, often close to zero. This indicates a poor alignment between automated and manual grading. A deeper analysis reveals that non-LLM-based models exhibit a uniform majority classification phenomenon, leading to skewed grading patterns. This suggests that LLM-based models provide more reliable grading results compared to their non-LLM counterparts. Comparing GPT-4o with prompt-optimized methods further highlights the importance of optimization. Optimized prompts consistently enhance grading performance while reducing variance across different questions. This finding confirms that directly applying raw human-provided rubrics is suboptimal, and prompt optimization is necessary to fully leverage LLMs in automatic grading. Lastly, \textit{GradeOpt} outperforms the state-of-the-art (SOTA) automatic prompt optimization method, APO, across all questions. This result demonstrates the superior effectiveness of \textit{GradeOpt} in improving grading performance, reinforcing its advantage over existing methods.

\vspace{-3mm}
\begin{table}[!btph]
\centering
\caption{Performance of \textit{GradeOpt} and Baseline Models on $\mathcal{D}_1$. The best performed model of each metric is marked with \textbf{bold}, the second best one is marked with \underline{underline}.}
\label{tab:result_pilot}
\resizebox{.475\textwidth}{!}{
\begin{tabular}{@{}cccccc@{}}
\toprule
\multicolumn{1}{c|}{\textbf{Question}} & \textbf{RoBERTa} & \textbf{SBERT} & \textbf{GPT-4o} & \textbf{APO} & \textbf{GradeOpt} \\ \midrule
\multicolumn{6}{c}{\textbf{Accuracy} (Acc)} \\ \midrule
\rowcolor[HTML]{EFEFEF}\multicolumn{1}{c|}{$Q_1$} & 0.80 & 0.61 & 0.85 & \underline{0.90} & \textbf{0.92} \\
\multicolumn{1}{c|}{$Q_2$} & 0.81 & 0.70 & 0.72 & \underline{0.89} & \textbf{0.91} \\
\rowcolor[HTML]{EFEFEF}\multicolumn{1}{c|}{$Q_3$} & 0.76 & 0.76 & 0.75 & \underline{0.80} & \textbf{0.86} \\
\multicolumn{1}{c|}{$Q_4$} & \textbf{0.79} & \underline{0.74} & 0.51 & 0.67 & 0.70 \\
\rowcolor[HTML]{EFEFEF}\multicolumn{1}{c|}{$Q_5$} & \underline{0.79} & 0.69 & 0.64 & \underline{0.79} & \textbf{0.80} \\
\multicolumn{1}{c|}{$Q_6$} & 0.49 & 0.76 & 0.70 & \underline{0.81} & \textbf{0.84} \\
\rowcolor[HTML]{EFEFEF}\multicolumn{1}{c|}{$Q_7$} & 0.55 & \underline{0.68} & 0.51 & \underline{0.68} & \textbf{0.73} \\
\multicolumn{1}{c|}{$Q_8$} & 0.66 & 0.62 & 0.66 & \underline{0.85} & \textbf{0.89} \\ \midrule
\multicolumn{6}{c}{\textbf{Cohen's Kappa} ($\kappa_c$)} \\ \midrule
\rowcolor[HTML]{EFEFEF}\multicolumn{1}{c|}{$Q_1$} & 0.65 & 0.32 & 0.76 & \underline{0.85} & \textbf{0.88} \\
\multicolumn{1}{c|}{$Q_2$} & 0.66 & 0.42 & 0.56 & \underline{0.80} & \textbf{0.85} \\
\rowcolor[HTML]{EFEFEF}\multicolumn{1}{c|}{$Q_3$} & 0.00 & 0.00 & 0.38 & \underline{0.51} & \textbf{0.68} \\
\multicolumn{1}{c|}{$Q_4$} & 0.00 & 0.00 & 0.09 & \underline{0.35} & \textbf{0.36} \\
\rowcolor[HTML]{EFEFEF}\multicolumn{1}{c|}{$Q_5$} & 0.58 & 0.44 & 0.30 & \underline{0.60} & \textbf{0.63} \\
\multicolumn{1}{c|}{$Q_6$} & 0.00 & 0.17 & 0.55 & \underline{0.69} & \textbf{0.70} \\
\rowcolor[HTML]{EFEFEF}\multicolumn{1}{c|}{$Q_7$} & 0.00 & 0.41 & 0.33 & \underline{0.50} & \textbf{0.52} \\
\multicolumn{1}{c|}{$Q_8$} & 0.37 & 0.29 & 0.48 & \underline{0.75} & \textbf{0.80} \\ \midrule
\multicolumn{6}{c}{\textbf{Quadratic Weighted Kappa} ($\kappa_w$)} \\ \midrule
\rowcolor[HTML]{EFEFEF}\multicolumn{1}{c|}{$Q_1$} & 0.70 & 0.31 & 0.82 & \underline{0.88} & \textbf{0.89} \\
\multicolumn{1}{c|}{$Q_2$} & 0.81 & 0.52 & 0.75 & \underline{0.93} & \textbf{0.94} \\
\rowcolor[HTML]{EFEFEF}\multicolumn{1}{c|}{$Q_3$} & 0.00 & 0.00 & 0.61 & \underline{0.67} & \textbf{0.76} \\
\multicolumn{1}{c|}{$Q_4$} & 0.00 & 0.00 & 0.24 & \underline{0.41} & \textbf{0.54} \\
\rowcolor[HTML]{EFEFEF}\multicolumn{1}{c|}{$Q_5$} & 0.59 & 0.32 & 0.56 & \underline{0.68} & \textbf{0.71} \\
\multicolumn{1}{c|}{$Q_6$} & 0.00 & 0.17 & 0.62 & \underline{0.77} & \textbf{0.80} \\
\rowcolor[HTML]{EFEFEF}\multicolumn{1}{c|}{$Q_7$} & 0.00 & 0.48 & 0.58 & \underline{0.64} & \textbf{0.70} \\
\multicolumn{1}{c|}{$Q_8$} & 0.44 & 0.33 & 0.68 & \underline{0.84} & \textbf{0.87} \\ \bottomrule
\end{tabular}}
\end{table}

\begin{table}[!btph]
\centering
\caption{Performance of \textit{GradeOpt} and Baseline Models on $\mathcal{D}_2$. The best performed model of each metric is marked with \textbf{bold}, the second best one is marked with \underline{underline}.}
\label{tab:result_interaction}
\resizebox{0.475\textwidth}{!}{
\begin{tabular}{@{}cccccc@{}}
\toprule
\multicolumn{1}{c|}{\textbf{Question}} & \textbf{RoBERTa} & \textbf{SBERT} & \textbf{GPT-4o} & \textbf{APO} & \textbf{GradeOPT} \\ \midrule
\multicolumn{6}{c}{\textbf{Accuracy} (Acc)} \\ \midrule
\rowcolor[HTML]{EFEFEF}\multicolumn{1}{c|}{$Q_1$} & 0.84 & 0.84 & 0.84 & \underline{0.96} & \textbf{0.98} \\
\multicolumn{1}{c|}{$Q_2$} & 0.57 & 0.55 & 0.76 & \underline{\textbf{0.86}} & \underline{\textbf{0.86}} \\
\rowcolor[HTML]{EFEFEF}\multicolumn{1}{c|}{$Q_3$} & 0.69 & 0.67 & 0.88 & \underline{\textbf{0.92}} & \underline{\textbf{0.92}} \\
\multicolumn{1}{c|}{$Q_4$} & 0.65 & 0.65 & 0.78 & \underline{0.80} & \textbf{0.82} \\
\rowcolor[HTML]{EFEFEF}\multicolumn{1}{c|}{$Q_5$} & \underline{0.94} & \underline{0.94} & \underline{0.94} & \underline{0.94} & \textbf{0.98} \\
\multicolumn{1}{c|}{$Q_6$} & \underline{\textbf{0.98}} & \underline{\textbf{0.98}} & 0.88 & 0.92 & 0.94 \\
\rowcolor[HTML]{EFEFEF}\multicolumn{1}{c|}{$Q_7$} & 0.98 & 0.98 & \underline{\textbf{1.00}} & \underline{\textbf{1.00}} & \underline{\textbf{1.00}} \\
\multicolumn{1}{c|}{$Q_8$} & \underline{\textbf{0.92}} & \underline{\textbf{0.92}} & 0.47 & 0.76 & 0.78 \\
\rowcolor[HTML]{EFEFEF}\multicolumn{1}{c|}{$Q_9$} & \underline{\textbf{0.98}} & \underline{\textbf{0.98}} & 0.96 & 0.96 & \underline{\textbf{0.98}} \\
\multicolumn{1}{c|}{$Q_{10}$} & \underline{\textbf{0.88}} & \underline{\textbf{0.88}} & 0.73 & 0.86 & \underline{\textbf{0.88}} \\
\rowcolor[HTML]{EFEFEF}\multicolumn{1}{c|}{$Q_{11}$} & 0.94 & 0.94 & 0.90 & \underline{\textbf{0.96}} & \underline{\textbf{0.96}} \\ \midrule
\multicolumn{6}{c}{\textbf{Cohen's Kappa} ($\kappa_c$)} \\ \midrule
\rowcolor[HTML]{EFEFEF}\multicolumn{1}{c|}{$Q_1$} & 0.00 & 0.00 & 0.55 & \underline{0.83} & \textbf{0.92} \\
\multicolumn{1}{c|}{$Q_2$} & 0.05 & 0.00 & 0.52 & 0.72 & 0.72 \\
\rowcolor[HTML]{EFEFEF}\multicolumn{1}{c|}{$Q_3$} & 0.08 & 0.00 & 0.74 & \underline{\textbf{0.81}} & \underline{\textbf{0.81}} \\
\multicolumn{1}{c|}{$Q_4$} & 0.00 & 0.00 & 0.59 & \underline{0.62} & \textbf{0.65} \\
\rowcolor[HTML]{EFEFEF}\multicolumn{1}{c|}{$Q_5$} & 0.00 & 0.00 & \underline{\textbf{0.64}} & \underline{\textbf{0.64}} & \underline{\textbf{0.64}} \\
\multicolumn{1}{c|}{$Q_6$} & 0.00 & 0.00 & 0.22 & \underline{0.31} & \textbf{0.38} \\
\rowcolor[HTML]{EFEFEF}\multicolumn{1}{c|}{$Q_7$} & 0.00 & 0.00 & \underline{\textbf{1.00}} & \underline{\textbf{1.00}} & \underline{\textbf{1.00}} \\
\multicolumn{1}{c|}{$Q_8$} & 0.00 & 0.00 & 0.10 & \underline{0.15} & \textbf{0.17} \\
\rowcolor[HTML]{EFEFEF}\multicolumn{1}{c|}{$Q_9$} & 0.00 & 0.00 & \underline{0.48} & \underline{0.48} & \textbf{0.66} \\
\multicolumn{1}{c|}{$Q_{10}$} & 0.00 & 0.00 & 0.34 & \underline{0.56} & \textbf{0.60} \\
\rowcolor[HTML]{EFEFEF}\multicolumn{1}{c|}{$Q_{11}$} & 0.00 & 0.00 & 0.50 & \underline{\textbf{0.73}} & \underline{\textbf{0.73}} \\ \bottomrule
\end{tabular}}
\end{table}

Table~\ref{tab:result_interaction} presents the performance of baseline models and \textit{GradeOpt} on dataset $\mathcal{D}_2$. Consistent with the findings in Table~\ref{tab:result_pilot}, LLM-based ASAG methods consistently outperform non-LLM-based models, reaffirming the advantages of leveraging LLMs for ASAG tasks. Moreover, \textit{GradeOpt} consistently achieves the best performance among all baselines in Table~\ref{tab:result_interaction}, demonstrating its effectiveness across different datasets. This confirms that \textit{GradeOpt} is a generalizable framework suitable for various ASAG tasks. Comparing model performance across the two datasets reveals an expected trend: grading responses related to teacher knowledge yields lower performance metrics. This aligns with our expectations, as evaluating pedagogical knowledge is inherently more complex, requiring  deeper expertise and more intricate logical reasoning. Beyond performance improvements, LLM-based ASAG methods enhance grading transparency and comprehensibility. Educators can easily interpret the LLM’s scoring rationale, which can be further broken down at the expectation level. Each key concept or criterion specified in the rubric or learning objectives is individually assessed, allowing educators to evaluate how well student responses align with specific learning goals. This level of explainability fosters greater trust in automated grading and facilitates more informed instructional decisions.

\vspace{-.25cm}

\subsection{Adaptation Results}

\begin{table}[!btph]
\centering
\caption{Result for $Q_1$ and $Q_2$ over new dataset $\mathcal{D}_1'$ before ($G_{opt}$) and after ($G_{opt}'$) test-time training. The average confidence indicator (CI) on Pilot Dataset $\mathcal{D}_1$ is $\mu=-0.2$ and questions with CI ($\zeta < -0.2$) are marked with OOD.}
\label{tab:robustness_table}
\resizebox{.475\textwidth}{!}{
\begin{tabular}{@{}cccccc@{}}
\toprule
\textbf{Question} & \textbf{Metric} &\ \ \ \ $\bold{G_0}$\ \ \ \ &\ \ \ \ $\bold{G_{opt}}$\ \ \ \ &\ \ \ \ $\bold{G_{opt}'}$\ \ \ \ &\ \ \ \ $\bold{\Delta}$\ \ \ \ \\ \midrule
\multicolumn{1}{c|}{\multirow{4}{*}{$Q_1$}} & \multicolumn{1}{c|}{Acc} & 0.67 & 0.7 & \multicolumn{1}{c|}{0.78} & +0.08 \\
\multicolumn{1}{c|}{} & \multicolumn{1}{c|}{Kappa} & 0.49 & 0.52 & \multicolumn{1}{c|}{0.64} & +0.12 \\
\multicolumn{1}{c|}{} & \multicolumn{1}{c|}{CI\ ($\zeta$)} & - & -0.22 & \multicolumn{1}{c|}{-0.17} & +0.05 \\
\multicolumn{1}{c|}{} & \multicolumn{1}{c|}{OOD} & - & \checkmark & \multicolumn{1}{c|}{$\times$} & - \\ \midrule
\multicolumn{1}{c|}{\multirow{4}{*}{$Q_2$}} & \multicolumn{1}{c|}{Acc} & 0.75 & 0.82 & \multicolumn{1}{c|}{-} & - \\
\multicolumn{1}{c|}{} & \multicolumn{1}{c|}{Kappa} & 0.59 & 0.7 & \multicolumn{1}{c|}{-} & - \\
\multicolumn{1}{c|}{} & \multicolumn{1}{c|}{CI\ ($\zeta$)} & - & -0.16 & \multicolumn{1}{c|}{-} & - \\
\multicolumn{1}{c|}{} & \multicolumn{1}{c|}{OOD} & - & $\times$ & \multicolumn{1}{c|}{-} & - \\ \bottomrule
\end{tabular}}
\end{table}

To address \textbf{RQ2}, we collect an external dataset, $\mathcal{D}_1'$, containing responses to two questions, $Q_1$ and $Q_2$, from new teachers across the nation. Specifically, we gather 1,352 responses for $Q_1$ and 1,364 responses for $Q_2$. From these, we randomly select 100 responses per question to serve as the test-time training dataset, while the remaining responses are used for evaluation. By applying the optimized guidelines $G_{opt}$ learned from $\mathcal{D}_1$ in Section~\ref{sec:pilot_result}, we explore its grading performance over $\mathcal{D}_1'$. In addition, if the $G_{opt}$ fails to pass the OOD test, where the confidence indicator $\zeta < \mu$, we will implement the test-time training with the train split of national dataset $\mathcal{D'}_{train}$. Then, the adapted guideline will be tested over the same $\mathcal{D'}_{test}$ again. Overall, the results are reported in Table~\ref{tab:robustness_table}. From the table, we can find that $Q_1$ is marked with OOD Flag as its confidence indicator $\zeta = -0.22 < -0.2 = \mu$. By comparing its performance between Table~\ref{tab:result_pilot} and Tabel~\ref{tab:robustness_table}, we can confirm that it suffers great performance drops. Meanwhile, the confidence indicator of $Q_2$, $\zeta = -0.16 > -0.2 = \mu$, and its performance gap between $\mathcal{D}_2$ and $\mathcal{D}_2'$ is relatively smaller. These two observations indicate that the proposed confidence indicator is a valid indicator for the OOD detection purpose. On the other hand, by comparing $G_{opt}$ with the raw guidelines provided by $G_0$, we find that $G_{opt}$ consistently outperforms $G_0$. This observation indicates that even the automatically optimized guideline suffers from the OOD issue, it is still better than the raw guideline. Finally, by calculating the performance change between the guideline before and after the test-time training, we can find that \textit{GradeOpt} is able to get adapted to the new examples with only limited available annotated examples. In addition, the performance of guidelines after the test-time training is restored back to the acceptable grading range, e.g., Kappa $>0.6$, which indicates that the test-time training is an effective solution to help a high-performed $G_{opt}$ to quickly get applied to different datasets.

\vspace{-.25cm}
\subsection{Ablation Studies}  \label{ablation}

\begin{figure}[!btph]
\centering
\begin{minipage}[b]{0.4\textwidth}
    \centering
    \includegraphics[width=\textwidth, alt={Side-by-side bars: misconfidence strategy accuracy vs. random.}]{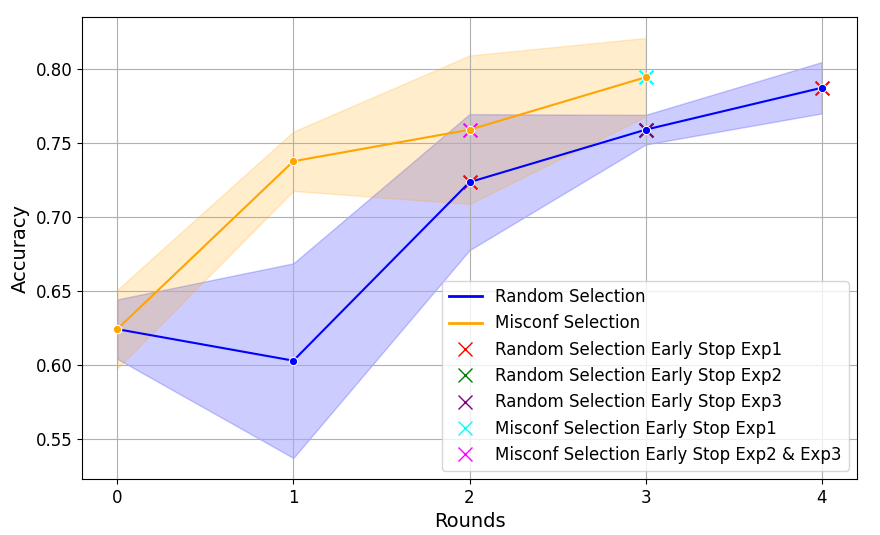} 
    (a) Accuracy Comparison
\end{minipage}
\hfill
\begin{minipage}[b]{0.4\textwidth}
    \centering
    \includegraphics[width=\textwidth, alt={Side-by-side bars: misconfidence strategy cohen's kappa vs. random.}]{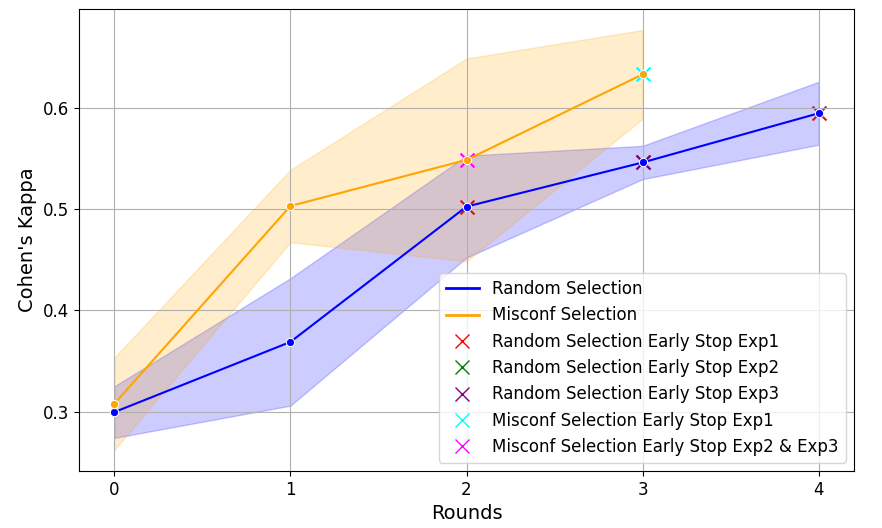} 
    (b) Cohen's Kappa Comparison
\end{minipage}
\caption{Performance comparison between \textit{GradeOpt} with \textit{misconfidence}-based and random-based outer batch selection strategies.}

\label{fig:misconf_ablation}
\end{figure}


To answer \textbf{RQ3}, we conduct ablation studies to the nested iteration introduced in Section~\ref{sec:training}. We choose to experiment with $Q_5$ from $\mathcal{D}_1$ as its relatively simple rubric design of $Q_5$ makes the shortest guideline prompt, leaving room for \textit{GradeOpt} to add in its reflective experience as it iteratively learns from $\mathcal{D}_2$. Thus, experimenting with $Q_5$ can better showcase \textit{GradeOpt}'s optimization power. With the experimental results shown in the following sections, we demonstrate the effectiveness of each component. 

\vspace{-0.15in}
\begin{figure}[!htbp]
\centering

\begin{minipage}[b]{0.33\textwidth}
    \centering
    \includegraphics[width=\textwidth, alt={Line plot: GradeOpt accuracy rises as outer-batch size increases,
       plateauing after size 8.}]{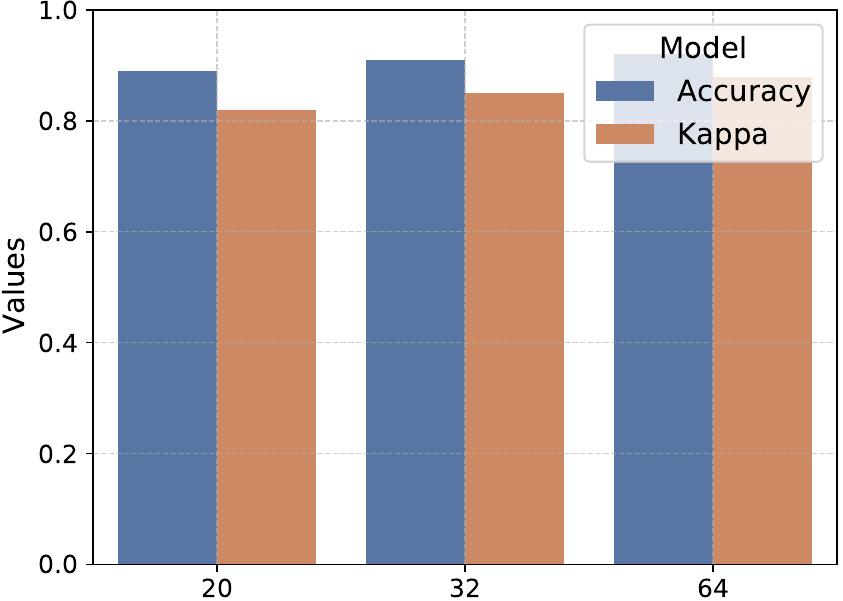} 
    (a) Outer Batch Size
    \label{fig:outer_batch}
\end{minipage}
\hfill
\begin{minipage}[b]{0.33\textwidth}
    \centering
    \includegraphics[width=\textwidth, alt={Line plot: GradeOpt accuracy rises as inner-batch size increases, plateauing after size 8.}]{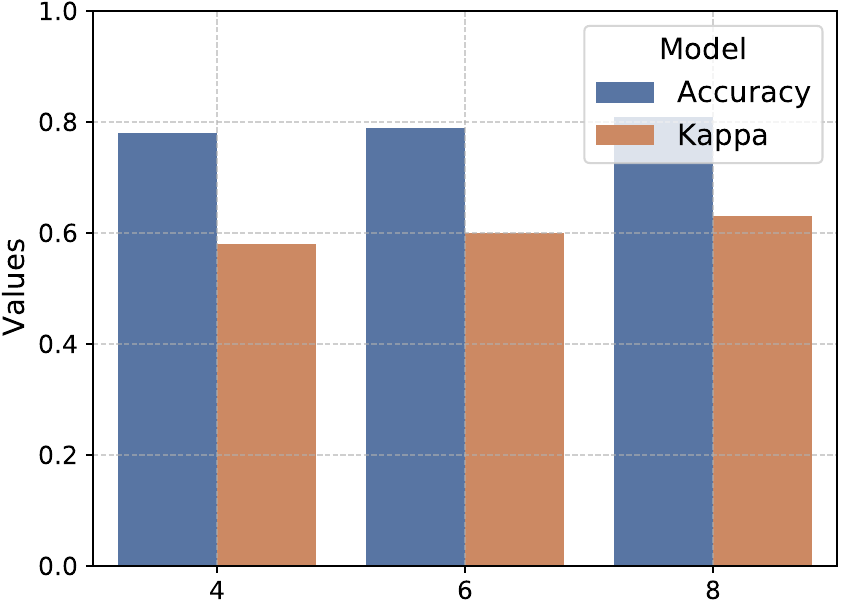} 
    (b) Inner Batch Size
    \label{fig:inner_batch}
\end{minipage}
\caption{Performance of \textit{GradeOpt} with different outer and inner iteration batch sizes.}
\label{fig:batch_size}
\end{figure}

\begin{figure}[!btph]
    \centering
    \includegraphics[width=.9\linewidth, alt={Influence of iteration number.}]{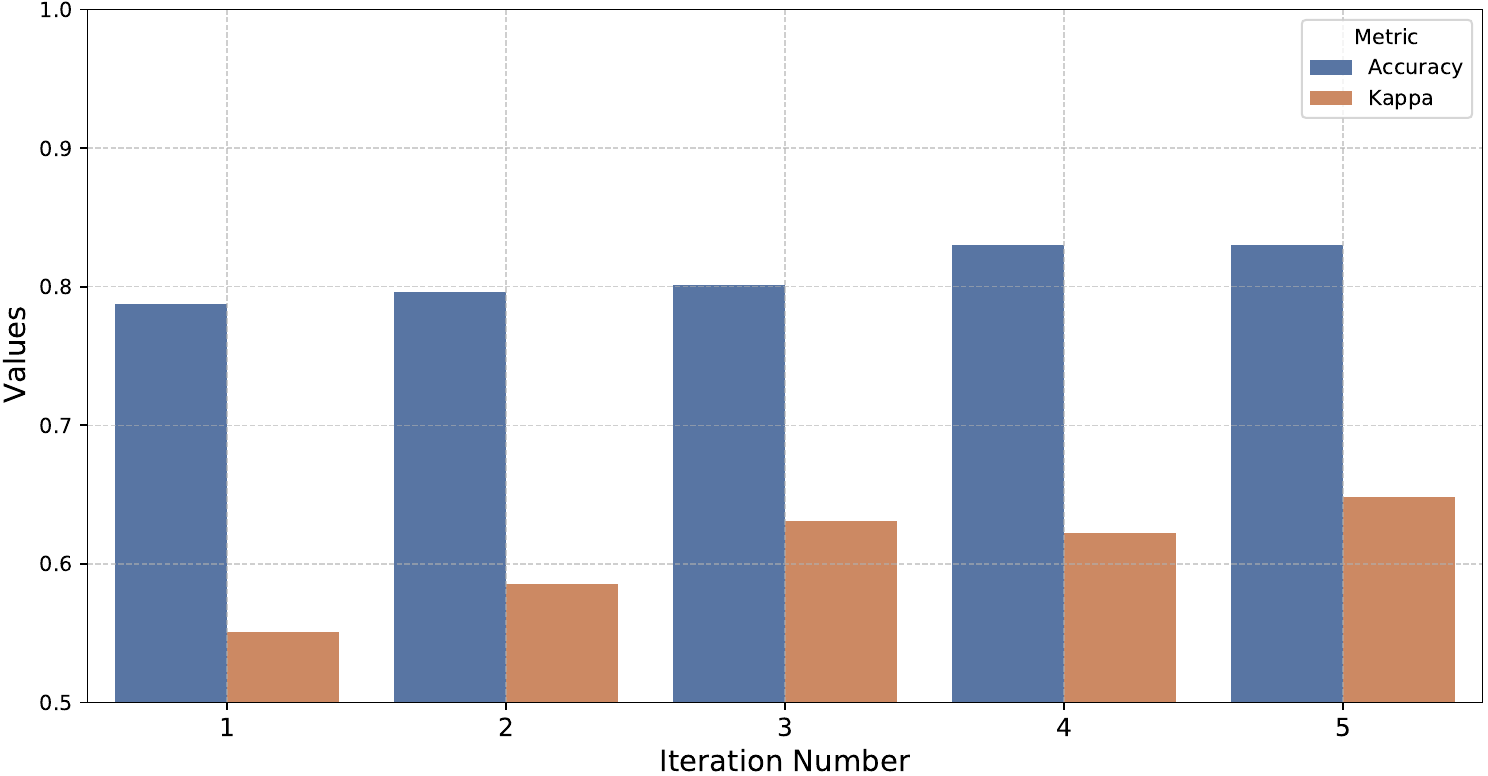}
    \caption{Performance of \textit{GradeOpt} with different numbers of iterations.}
    \label{fig:iteration_num}
\end{figure}

First, we demonstrate the superiority of our \textit{misconfidence}-based batch sampling strategy by comparing it with the random-based one. From Figure~\ref{fig:misconf_ablation},  we can observe that \textit{misconfidence}-based batch sampling results are more consistent and accurate. While random selection generates optimal guidelines in 2 to 4 rounds, \textit{misconfidence}-based selection consistently optimizes guidelines in 3 rounds. This, together with high predictive accuracy and alignment brought by \textit{misconfidence}-based selection, makes the system reliable in practical educational scenarios as the required training round number is coherent. Then, we conduct experiments over the sizes of outer batch $|b_{out}|\in\{20,32,64\}$ and inner batch $|b_{in}|\in\{4,6,8\}$, targeting at exploring the influence of those two hyper-parameters on the performance of \textit{GradeOpt}. From the results in Figure \ref{fig:outer_batch}, we observe an increasing trend in accuracy and kappa as its outer batch size increases. This observation suggests that increasing the number of examples is always beneficial to the final performance. Similarly, from the Figure~\ref{fig:inner_batch}, we find an consistent increasing trend of performance as error number increases. Based on these two findings, we can conclude that the larger batch size is likely to bring performance gains to the \textit{GradeOpt}. At last, we study how the iterations number impacts accuracy performance. In our experiment, we explore different iteration numbers, ranging from 1 to 5. Whilst increasing iteration on a minibatch, we utilize the early-stopping signal introduced in Section \ref{nested_iteration} to carefully monitor overfitting. As shown in Figure \ref{fig:iteration_num}, increasing iteration with the help of early-stopping signal can effectively lead to higher test accuracy as well as more stable performance. While five iterations produce higher accuracy, due to limited computational resource, we use three iterations as our default setting.

\section{Case Study}

In this section, we conduct case study by going through the refinement history and presenting the extraordinary performance of \textit{GradeOpt} in optimizing the guidelines to match with task specifications and training dataset. We showcase example responses of \textit{Reflector} and \textit{Refiner} separately. 

\subsection{Reflector} \label{reflector_case_study}

The \textit{Reflector} reflects and gives feedback when provided with a small selection of error samples. For example, as shown in Table \ref{tab:FA_reflector_case_study}, given a selection of errors in the prompt, \textit{Reflector} follows a two-step reflect-and-suggest approach. It first provides an explanation to each error, illustrating the reason for why the mistake is made and analyzing why the response should be assessed differently.
Based on this reflection, \textit{Reflector} next discusses its suggestion of rule modification, based on the evidence from error sample.

\renewcommand{\arraystretch}{1.3}
\begin{table}[]
\centering
\caption{Case Study for \textit{Reflector}}
\resizebox{.475\textwidth}{!}{
\begin{tabular}{p{0.13\textwidth}|p{0.4\textwidth}}
\toprule
\textbf{Example Error }& Teacher Response: ``They are equal proportions because they are situations during the same time and the same angle of the sun.'' 

- Expert Label: 2; AI Prediction: 1 \\\hline
\multirow{2}{*}{\begin{tabular}[c]{@{}c@{}}\textbf{Reflector's Error}\\ \textbf{Feedback}\end{tabular}}  & \textbf{Step 1: Explanation of Mistake} - The classifier predicted a score of 1 because the response mentions equal proportions and references conditions affecting proportionality (same time and same angle of the sun). However, it lacks specific evidence from the student's work and a thorough interpretation of the student's mathematical thinking. The correct label is 2 because the response does provide a clear explanation of the proportional relationship and the conditions under which it holds, which is a nuanced understanding of proportionality. \\\cline{2-2}
& \textbf{Step 2: Rule Modification} - Rule: If the response explains equal proportions and references conditions affecting proportionality, classify as 2, even if it lacks specific evidence from the student's work. \\
\bottomrule
\end{tabular}}

\label{tab:FA_reflector_case_study}
\end{table}

\subsection{Refiner} \label{refiner_case_study}

As introduced in Section \ref{refiner}, \textit{Refiner} is tasked with guideline modification given the feedback from \textit{Reflector}. \textit{Refiner} can edit by adding, changing, removing, etc. We first showcase an example of \textit{Refiner} adding a rule in Table ~\ref{tab:FA_refiner_case_study}, in response to \textit{Reflector}'s feedback in Section~\ref{reflector_case_study}. We observe that \textit{Refiner} takes \textit{Reflector}'s rules suggested in \textit{``Step 2: Rule Modification''}. The new rule explains two components for grading a certain type of answer: first, the answer pattern, by saying \textit{``if the teacher's response explains equal proportions and references conditions affecting proportionality, ..., even if it lacks specific evidence from the student's work''}; second, the score assignment, by saying \textit{``classify as 2''}.  It additionally adds in details from the example by noting \textit{``(e.g., same time, same angle of the sun)''} and includes the whole answer to give an elucidative grasping of to what answers this rule can be applied. To conclude, when adding rules \textit{Refiner} is to include new types of answers, concluding their patterns and explaining word or phrase details. Next, we show an example of \textit{Refiner} editing the guidelines. This happens when \textit{Refiner} determines that a defective rule can be adjusted to give a better explanation or more complete details. As the example in Table~\ref{tab:FA_refiner_case_study} shows, part of the given rule is misleading, so \textit{Refiner} revises this part. The changed rule first redefines the scenarios when this rule can be utilized, then it provides a detailed explanation by citing the answer and illustrating how the answer falls into the pattern category defined in the new rule to receive the correct score.

\begin{table}[]
\centering
\caption{Case Study for \textit{Refiner}}
\resizebox{.475\textwidth}{!}{
\begin{tabular}{>{\centering\arraybackslash}p{0.05\textwidth}|p{0.45\textwidth}}
\toprule

\textbf{Action} &\textbf{ \textit{Refiner}'s generated rules} \\\hline
Add & \textit{``1. Equal Proportions and Conditions:} \\

& \textit{- Rule: If the teacher's response explains equal proportions and references conditions affecting proportionality (e.g., same time, same angle of the sun), classify as 2, even if it lacks specific evidence from the student's work.} \\

& \textit{- Example: They are equal proportions because they are situations during the same time and the same angle of the sun.''} \\\hline

Edit & Before Edit: \textit{``- Award 1 point if ... explicit evidence. \uline{For instance, if the response mentions that the student understands the unit rate, which is related to the concept of proportionality.}''} (misleading statement) \\

& After Edit: \textit{``- Award 1 point if ... explicit evidence. \uline{For example, if the response states that the student might have a limited understanding of proportionality, it should be awarded 1 point. For instance, in the response ``their answer makes sense only if there is a proportional relationship between the height of the object and the length of the shadow," teacher mentioned ``limited/partial understanding of the proportional relationship" but lacked depth.}} (ambiguity resolved) \\
\bottomrule
\end{tabular}}

\label{tab:FA_refiner_case_study}
\end{table}

\vspace{-0.03in}

\section{Conclusion}

This paper explores fully automating guideline optimization to leverage LLM techniques including reflection and prompt engineering to solve ASAG tasks. We innovatively decompose the ASAG procedure into two steps: guideline optimization and grading. Specifically, we set our focus on automatic guideline optimization to avoid the manual efforts of composing a task-optimal guideline. To further prevent labeling a large amount of data, we propose a two-phase \textit{``train and test-adapt"} procedure to maximally tune a guideline on a small training set and securely ensure this optimized output is reliable for large-scale grading. The proposed \textit{GradeOpt} is a multi-agent guideline optimization system that iteratively leads the LLM to reflect on mistakes, learn answer patterns, and make improving modifications. Empirical experiments on two pedagogical datasets have demonstrated the effectiveness of \textit{GradeOpt}.

\vspace{-.25cm}
\section{Acknowledgments}
This work was supported in part by the National Science Foundation under Grant No. 1813760, No. 2405483, No. 2200757 and No. 2234015. Any opinions, findings, and conclusions or recommendations expressed in this material are those of the authors and do not necessarily reflect the views of the National Science Foundation. We thank Dr. Clare Carlson for help in revising one of the scoring rubrics.

%
\bibliographystyle{abbrv}
\bibliography{ref}  
\balancecolumns
\end{document}